\documentclass[11pt]{article}

\usepackage[utf8]{inputenc}
\usepackage[T1]{fontenc}
\usepackage[margin=1in]{geometry}
\usepackage{amsmath}
\usepackage{amssymb}
\usepackage{booktabs}
\usepackage{array}
\usepackage{hyperref}
\hypersetup{colorlinks=true, linkcolor=blue, citecolor=blue, urlcolor=blue}
\usepackage{enumitem}
\usepackage{graphicx}
\usepackage{tikz}
\usetikzlibrary{arrows.meta, positioning, shapes.geometric, fit, backgrounds, calc}
\usepackage{pgfplots}
\pgfplotsset{compat=1.18}
\usepackage{xcolor}
\usepackage{float}
\usepackage{mdframed}
\usepackage{needspace}
\mdfsetup{nobreak=true}
\usepackage{caption}
\usepackage{setspace}
\definecolor{geneblue}{RGB}{30,100,180}
\definecolor{envgreen}{RGB}{34,139,60}
\definecolor{signalred}{RGB}{190,40,40}
\definecolor{arrowgray}{RGB}{100,100,100}
\definecolor{boxbg}{RGB}{245,248,252}
\definecolor{warningbg}{RGB}{255,248,235}

\title{\textbf{Is It You or Your Environment?\\
\large A Bayesian Inference Framework for Genomically-Anchored\\
Personalized Physiological Interpretation}}

\author{Aruna Dey \quad Suraj Biswas\\
\small Dots-In\\
\small \texttt{cpo@dotsin.ai} \quad \texttt{ceo@dotsin.ai}}

\date{\today}

\begin{document}

\maketitle

\begin{abstract}
\noindent
Personalized health AI systems face a fundamental cold-start problem: machine learning models for physiological interpretation require weeks of individual behavioral data before they can distinguish constitutional variation from environmentally driven deviation. We propose a solution grounded in causal inference and Bayesian prior design. An individual's genomic profile serves as an \textbf{exogenous genetic anchor}---a domain-informed, personalized prior that is fixed at conception, immune to reverse causation, and available before a single behavioral observation is collected. Formally, the anchor initializes a Bayesian belief state over an individual's physiological set point $\hat{G} = \mu + \sum_i \beta_i g_i$, where $\beta_i$ are GWAS-derived effect sizes and $g_i \in \{0,1,2\}$ are risk-allele counts. Each incoming physiological measurement $P$ then produces a \emph{non-constitutional deviation} $\delta = P - \hat{G}$ whose distribution $\delta \sim \mathcal{N}(P - \mu_G,\, \sigma_G^2 + \sigma_\varepsilon^2)$ separates the signal attributable to environment and state from the constitutionally fixed baseline. As behavioral data accrue, the prior decays according to $\hat{G}_t = w(t)\hat{G}_{\text{genomic}} + [1-w(t)]\bar{P}_t$, transitioning from genome-dominated to empirical-baseline-dominated inference. The central AI contribution is a \emph{causal decomposition layer} that converts raw physiological streams into personalized, attribution-ready signals: the same observed HRV of 55\,ms generates a suppression hypothesis for a person whose prior predicts 80\,ms, and an enhancement hypothesis for a person whose prior predicts 30\,ms---a reversal impossible without a personalized anchor. We develop this architecture across six physiological domains, grading genomic priors by evidence strength and uncertainty ($\sigma_G^2$), distinguishing robustly replicated anchors (FTO, FADS1/2, FKBP5) from contested candidate genes (SLC6A4, MAOA, DRD2). We address the inference boundary between association, Mendelian randomization, and individual token causation, and define four constraints for deployment: evidence-graded priors, dynamic decay, ancestry-matched effect sizes, and attribution rather than deterministic output. The framework directly addresses the cold-start personalization problem in health AI without requiring historical behavioral data.
\end{abstract}

\section{Introduction: Is It You or Your Environment?}

A cardiologist seeing a patient's resting heart rate of 48 beats per minute asks, before anything else: is this person an athlete? The number alone means nothing. In an elite endurance runner it is healthy adaptation; in a sedentary sixty-year-old it triggers a workup for sick-sinus syndrome. The clinician resolves the ambiguity through context---occupational history, physical examination, longitudinal comparison---before forming a causal account of what is happening.

Continuous physiological monitoring from consumer sensors now generates millions of such numbers per person per year, but AI systems interpreting these streams face the same disambiguation problem without the clinician's contextual machinery. A heart-rate-variability reading of 55\,ms, a sleep midpoint of 2:30\,AM, a resting cortisol of 18\,nmol/L: each is a data point that requires a reference before it can carry meaning for inference. Two reference systems are currently used in health AI, and both are inadequate.

\paragraph{Population norms.}
Normative tables---``a healthy adult male aged 30--40 has a resting HRV of 40--80\,ms''---answer whether a reading is typical at the group level. They cannot say whether it is typical \emph{for this individual}. A person whose constitutional autonomic tone is near the 90th percentile will look ``normal'' on every population table even when their signal has been severely suppressed by an overtraining block, because the suppressed value still falls within the population range. The clinically important signal---``this is 30\% below my baseline''---is invisible to any population reference.

\paragraph{Personal behavioral baseline.}
A longitudinal personal baseline---the rolling average of a person's own prior readings---is far more informative, because it flags idiosyncratic deviations. But it does not exist at first contact, which leads to the cold-start problem formalized below.

\subsection{The Cold-Start Problem}
\label{sec:coldstart}

The cold-start problem is the central practical justification for this framework. A personal behavioral baseline---the most informative reference for physiological interpretation---requires approximately 7--30 days of consistent data per signal before it stabilizes \cite{jacobson2019}. During that window, a monitoring system operating on population norms cannot distinguish constitutional variation from non-constitutional deviation: the person whose HRV is constitutionally high looks the same as the person whose HRV is environmentally elevated. Every false attribution made in the first weeks of monitoring can seed incorrect causal hypotheses that persist long after the baseline has stabilized.

The cold-start problem is not merely a data-quantity problem. It is a \emph{reference-quality} problem: population norms are the wrong reference for an individual, and no amount of additional data collection fixes that. What is needed is a personalized reference that exists before any behavioral data are collected---a day-zero anchor.

An exogenous genetic anchor addresses this directly. Because the anchor is derived from germline sequence rather than behavioral observation, it is available on day one. It is not a perfect personalized reference---genetic contributions to most physiological traits are modest and noisy---but it is a principled one: the best available personalized estimate of where a physiological signal is likely to sit constitutionally, before any behavioral data are collected. As longitudinal data accumulate, the behavioral baseline progressively replaces the genetic anchor as the primary reference; the two are complementary across time, not competitors.

\paragraph{The proposal.}
We propose using an \textbf{exogenous genetic anchor} to bridge the cold-start gap and support causal interpretation throughout the monitoring lifecycle. The exogenous genetic anchor is defined as the genomic estimate of an individual's constitutional physiological set point, derived from GWAS effect sizes applied to the individual's genotype. It is \emph{exogenous} because genotype is fixed at conception and cannot be caused by downstream behavior, environment, or physiological state---the same property that licenses Mendelian randomization as a causal inference tool. Against this anchor, observed physiological signals decompose into constitutional expectation and \emph{non-constitutional deviation}---a signal that is, by construction, non-genetic, and therefore candidate-causal.

The framework's central claim is that physiological interpretation should be performed relative to individualized constitutional expectations rather than population norms. Two people with the same observed physiological value can require opposite causal interpretations once their constitutional expectations are known; population norms make this distinction impossible.

The contribution is a conceptual and methodological framework positioned at the intersection of Bayesian inference, causal AI, and genomics. We map six physiological domains to their genetic determinants, grade evidence rigorously, develop the causal machinery, and derive constraints for honest implementation. The framework's defining feature is \emph{calibrated restraint}: an exogenous genetic anchor is genuinely informative but genuinely weak, and the architecture must reflect both.

\vspace{4pt}
\noindent\textbf{Contributions to AI and machine learning.}
\begin{itemize}[noitemsep, topsep=3pt, leftmargin=*]
\item \textbf{Cold-start prior design:} A principled method for initializing personalized physiological inference before behavioral data exist, using exogenous genetic information as a domain-informed Bayesian prior.
\item \textbf{Causal decomposition layer:} A formal decomposition $\delta = P - \hat{G}$ that converts raw physiological streams into attribution-ready signals by removing the constitutionally fixed component, directly narrowing the causal search space for downstream reasoning.
\item \textbf{Uncertainty-gated inference:} An explicit uncertainty model $\hat{G} \sim \mathcal{N}(\mu_G, \sigma_G^2)$ where evidence grade propagates to inference confidence, preventing low-evidence genetic priors from driving spurious attributions.
\item \textbf{Dynamic belief updating:} A prior-decay architecture $\hat{G}_t = w(t)\hat{G}_{\text{genomic}} + [1-w(t)]\bar{P}_t$ that transitions from genome-dominated to empirical-baseline-dominated inference as data accrue---a principled solution to the cold-start-to-personalization transition.
\item \textbf{Honest causal ladder:} An explicit mapping of what the inference system delivers (Rung 1: ranked causal hypotheses) versus what requires experimental evidence (Rungs 2--3: intervention and counterfactual), preventing over-claiming in deployed health AI systems.
\end{itemize}

\section{Background}

\subsection{The Genetic Architecture of Physiological Traits}

The starting point is what genetics can and cannot tell us about physiology. Twin studies have established substantial heritability for many physiological traits: HRV heritability is estimated at 47--64\% \cite{molenaar2016}, body-mass-related traits at 40--70\% \cite{silventoinen2010}, and circadian chronotype at 40--50\%. These figures confirm that an exogenous genetic anchor is not vacuous---a meaningful portion of physiological variation is constitutionally fixed.

But heritability from twins is not the same as variance recoverable from measured common variants (SNPs). GWAS-based estimates of ``SNP heritability'' typically run at roughly half the twin-based figure \cite{powerpluess2015}: circadian chronotype, for example, shows twin heritability near 40--50\% but common-SNP heritability of only 14\% \cite{jones2019}. This gap---the ``missing heritability'' \cite{manolio2009}---reflects the contribution of rare variants, structural variation, dominance, and gene--environment interaction that SNP panels cannot capture. Furthermore, the trait architecture revealed by GWAS is overwhelmingly polygenic: signal is spread across hundreds to thousands of loci, each with tiny effect size, as formalised in the omnigenic model \cite{boyle2017}. A named single-gene panel therefore captures only a sliver of the measurable genetic variance.

The implication is direct: a genomic set-point estimate from a tractable variant panel is \emph{weak}. It shifts a prior modestly; it does not determine a value.

\subsection{The Candidate-Gene Crisis}

For decades, behavioral and physiological genomics sought to explain individual differences through named ``candidate genes'' selected for mechanistic plausibility---COMT for prefrontal dopamine, SLC6A4 for serotonin reuptake, DRD2 for reward. This program has largely failed replication. A landmark pre-registered analysis of 18 historically prominent candidate-gene hypotheses for depression found no support across multiple large samples \cite{border2019}. Meta-analyses of the 5-HTTLPR $\times$ stress interaction---once the most celebrated gene-by-environment finding in psychiatry---found no robust main effect or interaction upon aggregating large, well-powered studies \cite{risch2009, culverhouse2018}. Table~\ref{tab:candidate} details the replication status of the most commonly applied behavioral candidate genes, which the present framework uses as a \emph{cautionary tier} rather than as reliable anchors.

\subsection{What a Genetic Anchor Is and Is Not}

Given this landscape, what exactly does an exogenous genetic anchor contribute? It is not a prediction of an individual's value. It is a \emph{probabilistic prior over that individual's set point}, calibrated by the strength of the evidence for each domain and carried with uncertainty proportional to that strength. Its unique property---the one no behavioral reference shares---is exogeneity: the genome cannot be changed by downstream physiology, behavior, or environment, so an anchor derived from it is immune to the reverse-causation that contaminates all behavioral references.

\section{The Causal Framework}

\subsection{The Causal Graph}

\begin{figure}[t]
\centering
\includegraphics[width=\textwidth]{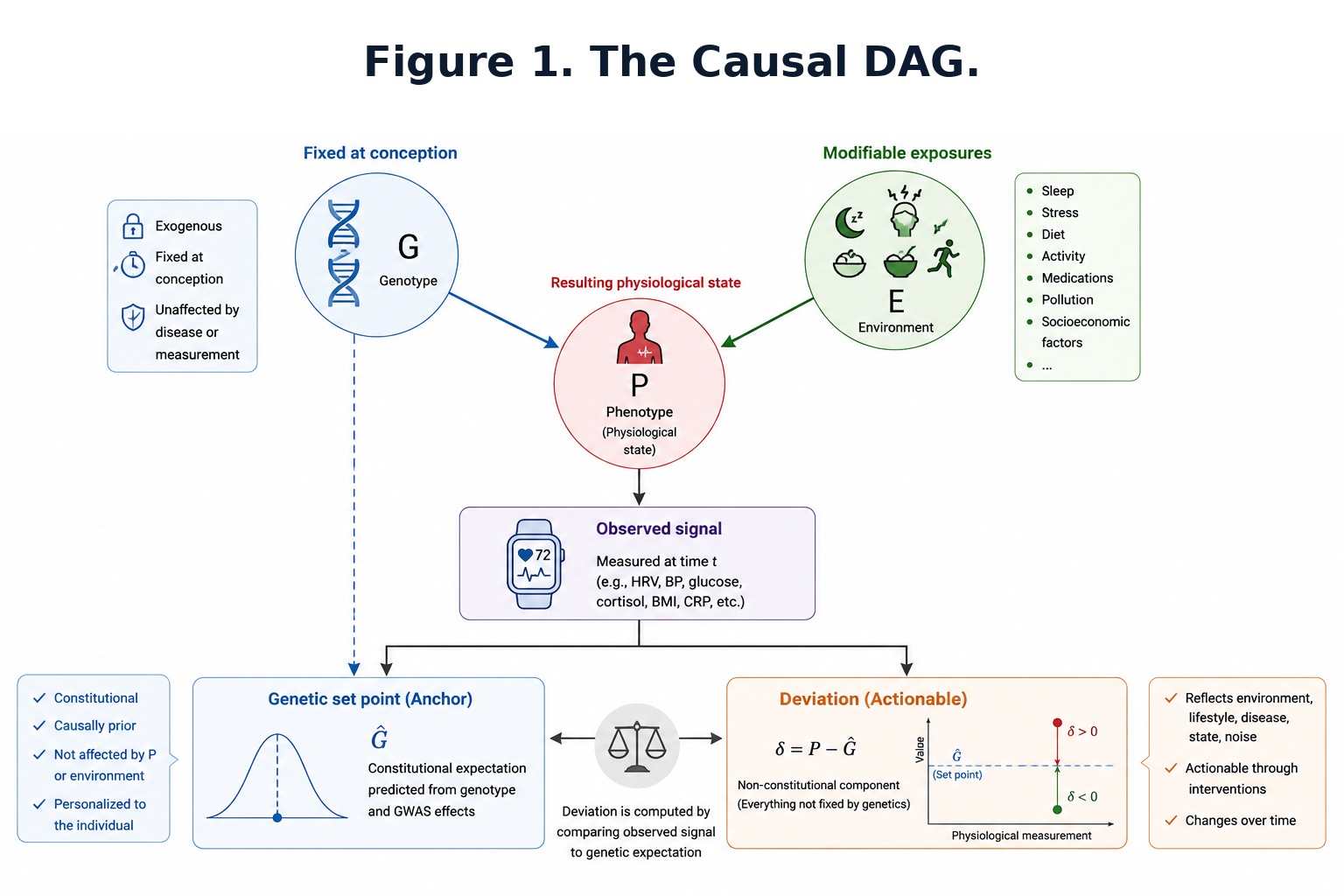}
\caption{Genotype $G$ is the root node (no incoming edges: it cannot be caused by downstream variables). Environment $E$ and $G$ together produce the phenotype $P$, which is what the sensor measures. Decomposing the observed signal into the genetic set point $\hat{G}$ and the deviation $P-\hat{G}$ isolates the non-constitutional, actionable component. Because $G$ is exogenous, the deviation cannot be an artifact of reverse causation from $P$ back to the anchor.}
\label{fig:dag}
\end{figure}

Figure~\ref{fig:dag} formalises the causal structure. Genotype $G$ sits at the root with no incoming edges, because nothing downstream---no behavior, no environment, no physiological state---can alter the germline sequence. This exogeneity is the same property that licenses Mendelian randomization \cite{daveysmith2003}: genetic variants serve as natural instrumental variables precisely because they are randomly allocated at conception. Environment $E$ and $G$ jointly produce the phenotype $P$, and measurement noise further separates $P$ from the sensor reading. Decomposing the reading into the genetic set point $\hat{G}$ and the deviation $P-\hat{G}$ isolates the non-genetic component, which is the one that environment, state, and behavior can modify.

\subsection{The Decomposition}

\begin{equation}
\label{eq:decomp}
P \;=\; \underbrace{G}_{\text{genetic set point}} + \underbrace{E}_{\text{environmental + state + noise}}
\qquad
\mathrm{Var}(P) = \mathrm{Var}(G) + \mathrm{Var}(E) + 2\,\mathrm{Cov}(G,E)
\end{equation}

Used as an interpretive device, equation~(\ref{eq:decomp}) motivates a single operation: estimate $\hat{G}$ from the genomic profile, compute the residual $\delta = P - \hat{G}$, and treat $\delta$ as the signal of interest. The deviation $\delta$ is the quantity that is candidate-causal and potentially actionable, because it is the portion of the reading that is not constitutionally fixed.

\subsection{Computing the Genetic Set Point}

The genetic set point $\hat{G}$ for a physiological signal is estimated as a weighted sum of per-locus contributions:

\begin{equation}
\label{eq:setpoint}
\hat{G} \;=\; \mu \;+\; \sum_{i=1}^{n} \beta_i \, g_i
\end{equation}

where $\mu$ is the population mean for the signal, $\beta_i$ is the GWAS-derived effect size (beta value) for locus $i$, and $g_i \in \{0, 1, 2\}$ is the individual's risk-allele count at that locus. This is the standard polygenic score estimator applied to a continuous physiological outcome rather than a disease endpoint.

\paragraph{Uncertainty in the set-point estimate.}
Because $\beta_i$ values are themselves estimated with error and the genetic contribution to most physiological traits is polygenic and partially unobserved, $\hat{G}$ is an uncertain estimate rather than a known value. We represent this uncertainty as:

\begin{equation}
\label{eq:uncertainty}
\hat{G} \;\sim\; \mathcal{N}\!\left(\mu_G,\; \sigma_G^2\right)
\end{equation}

where $\mu_G = \mu + \sum_i \beta_i g_i$ and $\sigma_G^2$ reflects three additive sources of variance: (i) sampling error in the $\beta_i$ estimates from the source GWAS; (ii) unobserved heritability not captured by the variant panel; and (iii) ancestry mismatch, which inflates variance when effect sizes are transferred across populations. In practice, $\sigma_G^2$ is larger for behavioral and neurocognitive traits (where candidate-gene effects are weak and polygenic) than for metabolic traits (where FTO and FADS loci carry stronger individual-level signal).

\paragraph{The deviation and its interpretation.}
Given this uncertainty, the deviation is itself a distribution rather than a point:

\begin{equation}
\label{eq:deviation}
\delta \;=\; P \;-\; \hat{G} \;\sim\; \mathcal{N}\!\left(P - \mu_G,\; \sigma_G^2 + \sigma_\varepsilon^2\right)
\end{equation}

where $\sigma_\varepsilon^2$ captures measurement noise in the observed signal $P$. A deviation is considered informative only when $|\delta|$ substantially exceeds $\sigma_G$ --- that is, when the observed signal lies well outside the uncertainty band of the genetic expectation. This uncertainty-gating is the operational equivalent of the convergence requirement discussed in Section~\ref{sec:constraints}: a single noisy reading against a weak genetic prior does not license a causal attribution.

\paragraph{Prior decay.}
As longitudinal behavioral data accrue, the empirical personal mean $\bar{P}$ converges to a stable individual baseline. The set-point estimate transitions from genomic-prior-dominated to empirical-baseline-dominated according to a decay schedule:

\begin{equation}
\label{eq:decay}
\hat{G}_t \;=\; w(t)\,\hat{G}_{\text{genomic}} \;+\; [1 - w(t)]\,\bar{P}_{t}
\end{equation}

where $w(t) \to 1$ at $t=0$ (cold-start) and $w(t) \to w_{\min} > 0$ as $t \to \infty$, ensuring the exogenous genetic anchor retains a non-zero interpretive role throughout. The specific form of $w(t)$ and the floor $w_{\min}$ are implementation decisions; the qualitative architecture is that the exogenous genetic anchor is most influential at cold-start and least influential once a long personal baseline exists.

\paragraph{Evidence grading propagates to uncertainty.}
Table~\ref{tab:map} grades the evidence for each domain. That grading is operationalized through $\sigma_G^2$: strong-evidence domains (FTO, FADS1/2) carry lower $\sigma_G^2$ and thus tighter set-point intervals; weak-evidence domains (COMT, DRD2, HTR2A) carry higher $\sigma_G^2$, widening the uncertainty band and requiring larger observed deviations before any attribution is generated.

\subsection{The ``Normal for Whom'' Reversal}

\begin{figure}[t]
\centering
\includegraphics[width=\textwidth]{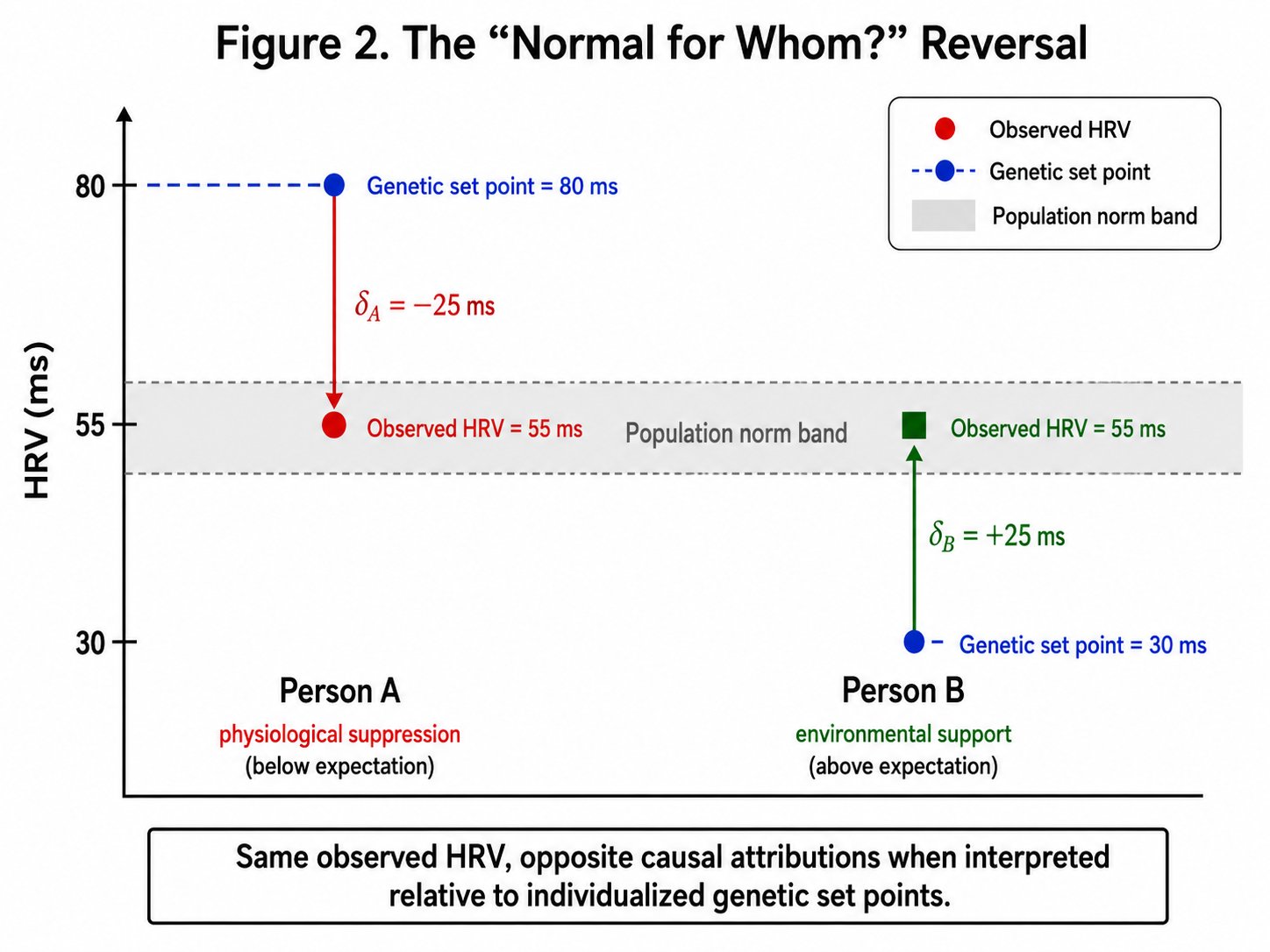}
\caption{Both persons A and B show an identical observed HRV of 55\,ms, which falls within the population norm band (shaded). Against their respective genetic set points, however, Person~A (high set point, 80\,ms) shows a negative deviation $\delta_A = -25$\,ms suggesting physiological suppression, while Person~B (low set point, 30\,ms) shows a positive deviation $\delta_B = +25$\,ms suggesting environmental support. Same observed value, opposite causal attributions---the reversal is only possible with an individualized genetic reference.}
\label{fig:reversal}
\end{figure}

Figure~\ref{fig:reversal} makes the reversal concrete. Person A carries a genotype associated with a high autonomic set point (approximated at 80\,ms). Their observed HRV of 55\,ms lies within the population norm band but sits 25\,ms \emph{below} their genetic expectation. The deviation $\delta_A = -25$\,ms is the signal: something is suppressing their autonomic tone. Plausible candidates include accumulated sleep debt, a heavy training load, or elevated chronic stress. Person B carries a genotype associated with a low set point (30\,ms). The same 55\,ms reading now sits 25\,ms \emph{above} expectation: their environment is currently supporting their autonomic function better than their constitution predicts. The population norm treats both as ``normal.'' The genetic-baseline framework treats them as physiologically opposite situations requiring opposite causal questions.

\section{Genetic Set Points Across Six Physiological Domains}

The decomposition's value depends entirely on the quality of $\hat{G}$, which varies by domain. Table~\ref{tab:map} summarises representative genes, mechanisms, and evidence grades for six physiological domains; Table~\ref{tab:candidate} presents the cautionary tier. We discuss each domain in evidence order---strongest anchors first---and use the three strongest (FTO, FKBP5, chronotype) as worked examples of the full causal-attribution workflow. The autonomic domain (HRV) is treated separately as a focused illustration of the ``normal for whom'' reversal mechanism rather than as a reliable set-point anchor, because the genetic contribution to HRV is genuine but small, polygenic, and context-dependent in ways that limit its standalone interpretive value.

\begin{table}[t]
\centering
\scriptsize
\renewcommand{\arraystretch}{1.6}
\begin{tabular}{>{\raggedright}p{2.0cm} >{\raggedright}p{2.2cm} >{\raggedright}p{2.8cm} >{\raggedright}p{5.2cm} >{\raggedright\arraybackslash}p{1.8cm}}
\toprule
\textbf{Domain} & \textbf{Key gene(s)} & \textbf{Physiological set point} & \textbf{Passive behavioral signals} & \textbf{Evidence} \\
\midrule
Metabolic / appetite &
\emph{FTO} (rs9939609) &
Satiety threshold; resting metabolic rate tendency &
Meal-timing regularity; post-meal step count; snacking frequency; energy-level proxy via typing speed across the day &
\textbf{Strong} \\

Fatty-acid / inflammatory &
\emph{FADS1/2} &
Constitutive PUFA ratio; systemic inflammatory tone &
Post-exertion HRV recovery time; resting HRV floor; energy variability across days; mood instability from communication tone shifts &
\textbf{Strong} \\

Stress-axis / cortisol &
\emph{FKBP5} (rs1360780) &
HPA feedback speed; cortisol-recovery ceiling &
Communication-frequency drop post-stressor; sleep fragmentation; typing speed under load; EDA; elevated resting HR duration &
\textbf{Moderate-- strong} \\

Autonomic tone &
Polygenic (\emph{GNG11, RGS6, HCN4}); \emph{COMT} minor nudge &
Resting HRV; resting heart rate &
Wearable RMSSD at rest; resting HR; HRV response to training; recovery HRV after sleep &
\textbf{Moderate, polygenic} \\

Circadian / chronotype &
Polygenic (\emph{PER1/2/3, CRY1, ARNTL}); rare \emph{CSNK1D, CRY2} &
Constitutional sleep midpoint; circadian phase &
First phone-pickup timestamp; sleep-onset from screen-off; activity-onset from step count; message send-time distribution; typing-speed rhythm &
\textbf{Moderate, polygenic} \\

Dopaminergic / serotonergic (cautionary) &
\emph{COMT, DRD2, SLC6A4} &
Prefrontal dopamine tone; reward sensitivity; serotonin reuptake &
Typing speed and error-rate ratio; communication latency; screen-wakeup frequency; app-diversity novelty index &
\textbf{Weak / contested} \\
\bottomrule
\end{tabular}
\caption{Three-Layer Framework: Genomic Anchor $\to$ Physiological Set Point $\to$ Passive Behavioral Signal. Each row maps a genomic anchor to the physiological set point it shifts and to the passive behavioral signals whose deviations form the actionable inference layer. Evidence grade reflects replication breadth and effect-size robustness.}
\label{tab:map}
\end{table}

\begin{figure}[t]
\centering
\includegraphics[width=\textwidth]{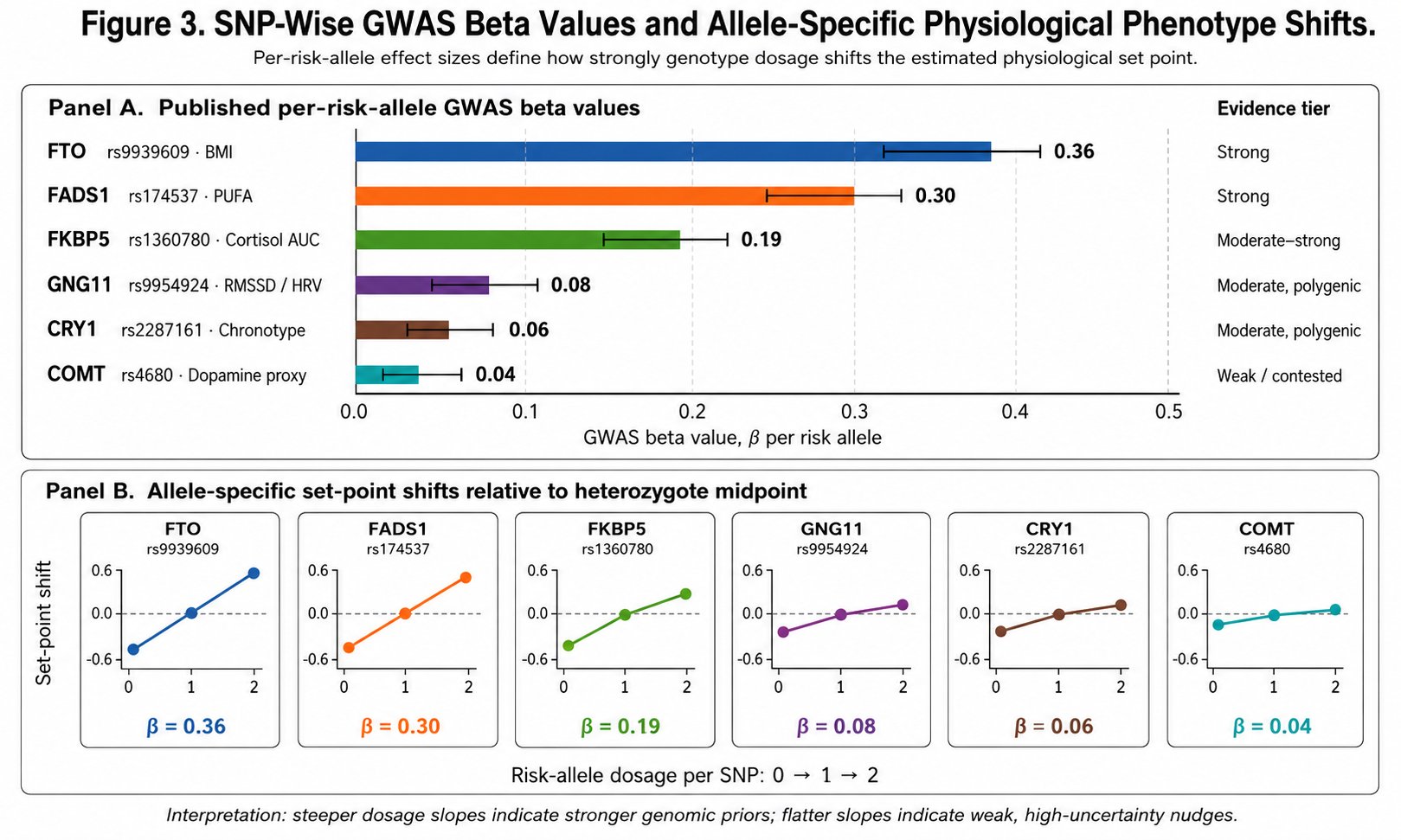}
\caption{\textbf{Panel~A:} Published per-risk-allele GWAS effect sizes with 95\% CI bars for six SNPs across five physiological domains.
FTO rs9939609 $\to$ BMI: $\beta \approx 0.36$\,kg/m$^2$ \cite{frayling2007,speliotes2010};
FADS1 rs174537 $\to$ circulating PUFA ratio: $\beta \approx 0.30$\,SD \cite{tanaka2009};
FKBP5 rs1360780 $\to$ cortisol AUC: $\beta \approx 0.19$\,SD \cite{klengel2013};
GNG11 rs9954924 $\to$ RMSSD (HRV): $\beta \approx 0.08$\,SD \cite{nolte2017};
CRY1 rs2287161 $\to$ chronotype: $\beta \approx 0.06$\,SD \cite{jones2019};
COMT rs4680 $\to$ prefrontal dopamine proxy: $\beta \approx 0.04$\,SD \cite{egan2001}.
Strong metabolic anchors carry 4--9$\times$ larger effect sizes than the dopaminergic candidate gene.
\textbf{Panel~B:} Allele-dosage set-point shifts (risk-allele count 0$\to$1$\to$2) relative to the heterozygote midpoint for each SNP. Steeper slopes (FTO, FADS1, FKBP5) indicate stronger per-allele constitutional shifts and lower $\sigma_G^2$; flatter slopes (HRV loci, chronotype, COMT) indicate weak, high-uncertainty nudges. \emph{Beta values are from published sources as cited; Panel~B slopes are directional approximations scaled to published effect sizes.}}
\label{fig:snp_betas}
\end{figure}

\subsection{Primary Example: FTO and the Metabolic Set Point}

Among all domains in Table~\ref{tab:map}, the metabolic domain offers the strongest, most practically consequential genomic set point. FTO rs9939609 is the most replicated common variant for body mass in the human genome, the gene-to-physiology chain is mechanistically characterized, and the actionable consequence of misattributing a constitutional tendency as a behavioral failing is directly harmful. This makes FTO the natural flagship illustration of the framework's value.

\begin{mdframed}[backgroundcolor=boxbg, linecolor=geneblue, linewidth=1pt]
\textbf{Worked Example --- Metabolic set point.} A person carries two copies of the FTO risk allele (rs9939609 A/A). Their exogenous genetic anchor shifts their expected satiety set point downward---a constitutional tendency toward positive energy balance, not a failure of will. Without this prior, continuous data showing evening snacking, elevated resting metabolic signals, and slow satiety responses would generate a false behavioral attribution: the system would rank stress or poor habits as the top causal candidates. With the exogenous genetic anchor, the deviation from the expected metabolic baseline is much smaller, and the ranked hypothesis correctly identifies a constitutional low-satiety signal amplified by an obesogenic food environment. The actionable question changes from ``why can't this person control eating?'' to ``what environmental modifications---food timing, meal composition, satiety-enhancing interventions---counteract a constitutional predisposition?'' The exogenous genetic anchor has converted a description into a correctly attributed causal hypothesis.
\end{mdframed}

The \emph{FTO} locus (rs9939609) is among the most robustly replicated common-variant associations in human genetics \cite{frayling2007}. The risk allele acts not through FTO protein itself but through cis-regulatory effects on \emph{IRX3} and \emph{IRX5} in hypothalamic circuits governing appetite and satiety: obesity-associated intronic variants form long-range chromatin contacts with the \emph{IRX3} promoter, and disrupting this interaction in animal models reduces adiposity \cite{smemo2014}. Effect size per risk allele is approximately 0.3--0.4\,kg/m$^2$ in European-ancestry samples---small but among the largest single-variant effects known for a complex trait \cite{speliotes2010}. Risk-allele carriers show reduced satiety signalling constitutionally. The set-point interpretation: the exogenous genetic anchor shifts $\mu_G$ upward for appetite-related signals, and $\sigma_G^2$ for this domain is lower than for behavioral or neurocognitive traits, because the FTO-to-appetite chain is mechanistically validated and replication is broad.

\subsection{Second Strong Anchor: FADS1/FADS2 and Inflammatory Status}

Variation at \emph{FADS1} and \emph{FADS2} governs the efficiency of desaturase enzymes that convert dietary precursor fatty acids (linoleic acid, $\alpha$-linolenic acid) to long-chain polyunsaturated fatty acids (arachidonic acid, DHA)---substrates for eicosanoid signalling, neuronal membrane composition, and inflammatory resolution \cite{tanaka2009, calder2010}. Low-efficiency variants shift the individual's constitutive circulating fatty-acid profile toward a more pro-inflammatory state, independent of diet \cite{vaittinen2022}. This is a genuine set-point anchor: two people on identical diets can show different systemic inflammatory baselines solely because their desaturase conversion efficiency differs genetically. The deviation from this set point captures the environment-driven component of inflammatory tone, making FADS1/2 the strongest exogenous genetic anchor available for inflammation-related physiological signals.

\subsection{Second Main Example: FKBP5 and the Stress-Recovery Set Point}

FKBP5 is the most mechanistically complete example in the framework, because it demonstrates something FTO cannot: that the genomic set point interacts with environmental history through epigenetics, and that the same genotype can produce different effective set points in different people depending on their early-life experience. This makes FKBP5 both a strong anchor and the clearest illustration of why the framework must be dynamic.

\begin{figure}[t]
\centering
\includegraphics[width=\textwidth]{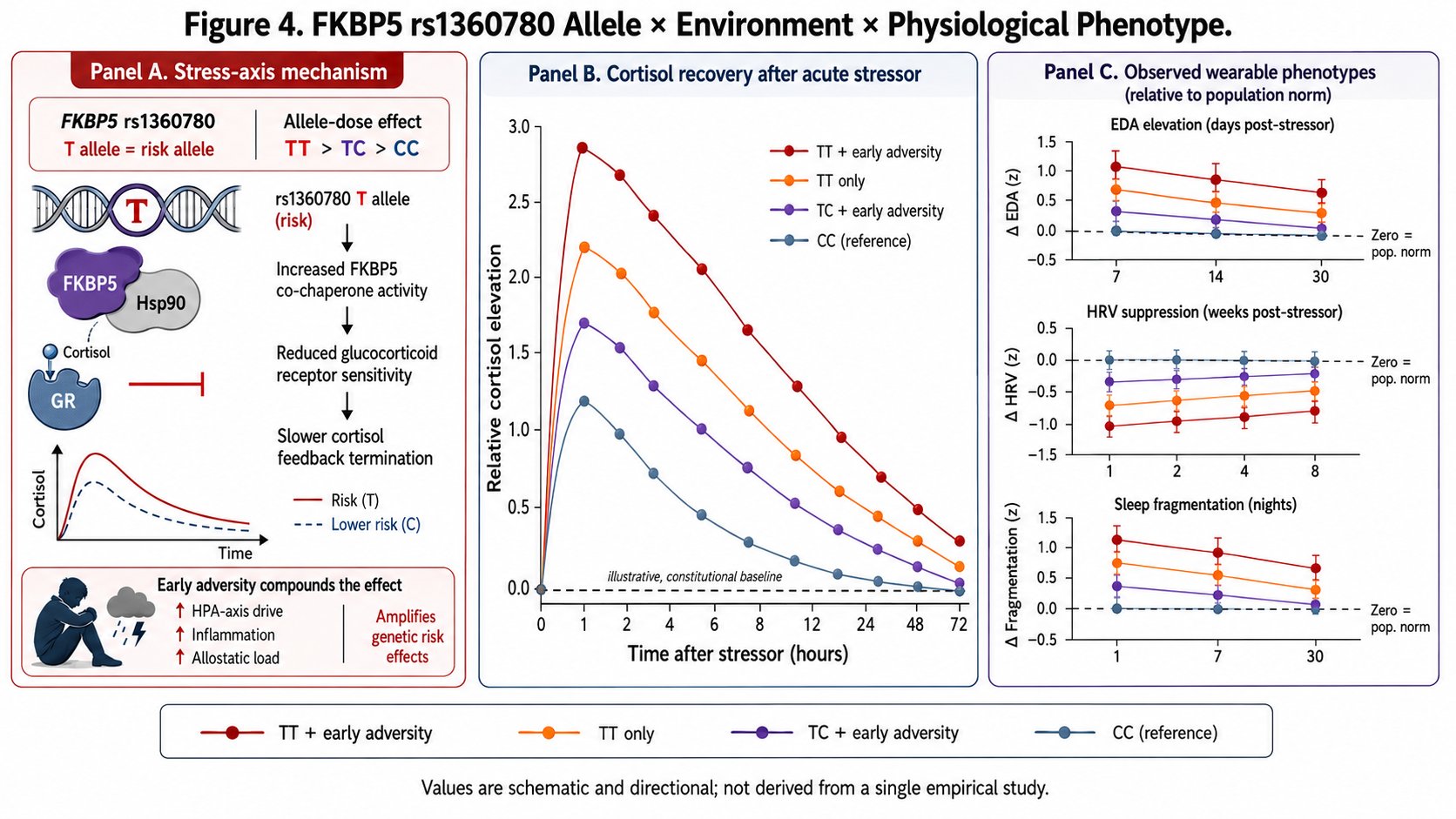}
\caption{\textbf{Panel A:} The rs1360780 T-allele (risk) increases FKBP5 co-chaperone activity, reducing glucocorticoid receptor sensitivity and slowing cortisol feedback termination; the effect is allele-dose-dependent (TT $>$ TC $>$ CC). Early-life adversity amplifies genetic risk effects through ↑ HPA-axis drive, ↑ inflammation, and ↑ allostatic load.
\textbf{Panel B:} Schematic cortisol recovery curves following an acute stressor, by genotype and early-adversity history (illustrative, based on published directional findings \cite{klengel2013,binder2009}). TT carriers with early adversity show the slowest recovery; CC carriers (reference) recover fastest regardless of adversity history.
\textbf{Panel C:} Corresponding wearable-observable phenotypes (EDA elevation, HRV suppression, sleep fragmentation) relative to population norm across days to weeks post-stressor. The exogenous genetic anchor lets the system attribute prolonged post-stressor signal elevation to the constitutional stress-recovery set point rather than generating a false hypothesis of a new ongoing stressor.
\emph{Values are schematic and directional; not derived from a single empirical study.}}
\label{fig:fkbp5}
\end{figure}

\begin{mdframed}[backgroundcolor=boxbg, linecolor=envgreen, linewidth=1pt]
\textbf{Worked Example --- Stress-axis set point.} A person carries the FKBP5 risk allele (rs1360780 T carrier) and experienced significant early-life adversity. Their exogenous genetic anchor---adjusted for the known GxE interaction---predicts a constitutionally slower cortisol recovery ceiling: even under normal conditions, their HPA axis runs closer to its stress-response threshold. Following a period of heavy work stress, continuous data show persistently elevated electrodermal arousal, disrupted sleep architecture, and blunted morning cortisol recovery weeks after the stressor has objectively passed. Without the exogenous genetic anchor, these deviations would generate two plausible false hypotheses: (1) a new unidentified stressor, or (2) poor sleep hygiene driving the EDA elevation. With the FKBP5 prior, the correct attribution emerges immediately: the person's constitutional cortisol-feedback mechanism is slow, so stress recovery takes longer than population norms predict. The ranked causal hypothesis is prolonged HPA dysregulation consistent with their genomic set point, not a new environmental cause. The n-of-1 test: structured recovery periods of known duration, measuring cortisol return-to-baseline against the genomic expectation rather than a population norm.
\end{mdframed}

\emph{FKBP5} encodes a glucocorticoid receptor co-chaperone that regulates HPA stress-axis sensitivity and the speed of cortisol feedback termination. By binding to the glucocorticoid receptor complex in the cytoplasm, FKBP5 reduces receptor sensitivity and delays nuclear translocation, forming an ultra-short intracellular negative feedback loop for the cortisol response \cite{binder2009}. Risk variants (rs1360780) are associated with slower cortisol recovery and, critically, with a gene-by-environment interaction: early-life adversity induces allele-specific demethylation of the \emph{FKBP5} gene in risk carriers, producing a persistent epigenetic alteration that prolongs HPA activation throughout life \cite{klengel2013}. This makes FKBP5 the clearest example of why the genomic set point is \emph{not} fully captured by genotype alone---epigenetic history is part of the anchor---and why identical genotypes can produce different effective set points across individuals.

\subsection{HRV as a Reversal Illustration: Why Autonomic Tone Needs a Genetic Reference}

The autonomic domain is not a strong set-point anchor---HRV is massively state-dependent, and the genetic contribution, while real (twin heritability 47--64\% \cite{molenaar2016}), is spread across dozens of sinoatrial-node loci that together explain only 0.9--2.6\% of variance in the largest GWAS \cite{nolte2017}. We include it here not as a flagship example but as the most vivid illustration of the ``normal for whom'' reversal, because HRV is the signal most readers will have seen on a consumer wearable and most often misinterpreted against population norms.

The reversal in Figure~\ref{fig:reversal} makes the point precisely: two individuals with the same observed HRV of 55\,ms require opposite causal interpretations once their genomic set points are known. For the person whose prior implies a high autonomic set point, the reading is a suppression signal; for the person whose prior implies a low one, it is evidence of environmental support. Population norms, by treating both as ``normal,'' miss both stories. The role of the exogenous genetic anchor in this domain is therefore primarily to \emph{personalize the interpretation threshold}, not to supply a confident prediction.

COMT (Val158Met) contributes modestly and inconsistently to this domain through a central, top-down pathway (prefrontal dopamine clearance $\to$ descending autonomic regulation $\to$ vagal tone), with an age-by-genotype interaction that reverses direction \cite{chang2019}. Its appropriate role is in the cautionary tier (Table~\ref{tab:candidate})---useful for understanding why two people with the same raw HRV might respond differently to stress, but unsuitable as a standalone set-point anchor.\footnote{COMT mechanistic detail: COMT uses S-adenosylmethionine (SAM) as a methyl donor to inactivate catecholamines via O-methylation. In the prefrontal cortex, where dopamine transporter expression is low, enzymatic breakdown dominates dopamine clearance. The rs4680 G$\to$A substitution changes valine to methionine at position 158, reducing thermal stability; the Met enzyme runs at roughly one-third to one-quarter of Val enzyme activity. Because alleles are co-dominant, Val/Met has intermediate activity---Val/Val $>$ Val/Met $>$ Met/Met in a roughly linear dosage relationship.}

\subsection{Third Main Example: Chronotype and Circadian Misalignment}

Chronotype is the most socially consequential set-point domain in the framework: a constitutionally delayed sleep phase, in a society built around conventional work schedules, generates a permanent gap between the genomic set point and the forced behavioral schedule. That gap is itself a deviation --- one caused not by a modifiable environmental factor but by structural social misalignment with a constitutional trait. This makes chronotype a uniquely policy-relevant anchor: the intervention is not behavior change but schedule redesign.

\begin{mdframed}[backgroundcolor=boxbg, linecolor=arrowgray, linewidth=1pt]
\textbf{Worked Example --- Chronotype set point.} A person's polygenic chronotype score places them in the delayed (evening) quartile. Their continuous sleep data during a conventional 9-to-5 work schedule shows a sleep-onset time of 1:30\,AM and alarm-driven waking at 7:00\,AM. The exogenous genetic anchor establishes that their constitutional sleep midpoint is approximately 3:00\,AM. The deviation between constitutional midpoint and social-schedule midpoint is 2--3 hours of chronic circadian misalignment---not poor sleep hygiene, not screen use, not caffeine. Without the exogenous genetic anchor, daytime fatigue, mood instability, and cognitive sluggishness would be attributed to modifiable behavioral causes. With it, the ranked causal hypothesis is structural misalignment between a constitutional trait and a social schedule, and the actionable question becomes: can schedule flexibility reduce this gap? The n-of-1 test: shift work schedule two hours later for two weeks and measure the deviation against the genomic chronotype expectation.
\end{mdframed}

Chronotype is a tractable polygenic anchor with a well-characterized GWAS landscape. The largest study---697,828 individuals---identified 351 loci including core clock genes (\emph{PER1}, \emph{PER2}, \emph{PER3}, \emph{CRY1}, \emph{ARNTL}), confirming pathway specificity, and validated associations against objective actigraphy \cite{jones2019}. Common-SNP heritability is 13.7\%---toward the lower end of twin estimates (40--50\%), illustrating the missing-heritability gap even in a well-powered study. The 5\% of individuals carrying the most morningness alleles differ in objective sleep timing by only 25 minutes from the 5\% carrying the fewest, so the set-point signal is real but modest. Rare Mendelian variants (\emph{CSNK1D}-T44A, \emph{CRY2}-A260T) are powerful in specific families but shrink dramatically in the general population---the same \emph{PER3} variant advancing sleep by 4 hours in a pedigree shifted timing by only 8 minutes at the population level \cite{jones2019}. The polygenic aggregate is the honest anchor for typical individuals.

\subsection{Dopaminergic and Serotonergic Signals: The Cautionary Tier}

The dopaminergic and serotonergic genes below are the most widely applied in consumer genomics panels, yet all belong in a cautionary tier rather than as set-point anchors. Their failure in large-scale replication is the expected consequence of applying single-gene thinking to traits with massively distributed genetic architecture \cite{boyle2017}. Table~\ref{tab:candidate} summarises their replication status.

\begin{table}[h!]
\centering
\small
\renewcommand{\arraystretch}{1.4}
\begin{tabular}{>{\raggedright}p{2.4cm} >{\raggedright}p{4.6cm} >{\raggedright\arraybackslash}p{5.4cm}}
\toprule
\textbf{Gene (variant)} & \textbf{Claimed association} & \textbf{Replication / evidence status} \\
\midrule
\emph{COMT} (Val158Met) & Prefrontal dopamine shaping executive function and stress response & Enzyme-activity effect robust; behavioral effects small, context-dependent, and reversible with age or stress \cite{cools2011, chang2019} \\
\emph{SLC6A4} (5-HTTLPR) & S-allele raising anxiety and depression under stress & Large pre-registered replications found no robust main effect or gene$\times$environment interaction \cite{risch2009, culverhouse2018, border2019} \\
\emph{MAOA} (uVNTR) & Low-activity allele linked to impulsivity and aggression (``warrior gene'') & Small, inconsistent effects; documented history of harmful misuse in forensic contexts \cite{richardson2011}; \textbf{unsuitable for individual risk attribution} \\
\emph{DRD2} (TaqIA) & A1 allele reducing D2 density, raising addiction risk (``reward deficiency'') & Density effect debated; behavioral associations modest and contested in meta-analysis \cite{jung2019} \\
\emph{DRD4} (7R VNTR) & 7R allele raising novelty-seeking and ADHD risk & Mixed in meta-analyses \cite{munoz2003}; novelty-seeking effect small; ADHD is highly polygenic \\
\emph{DRD3} (Ser9Gly) & Gly allele altering D3 affinity, raising impulsivity & Weak and inconsistent \cite{chukwueke2020} \\
\bottomrule
\end{tabular}
\caption{The Cautionary Candidate-Gene Tier. These associations have supporting literature but have failed rigorous large-scale replication, making them unreliable set-point anchors. They are included to illustrate the risk of applying the framework with overconfident priors.}
\label{tab:candidate}
\end{table}

\subsection{The Cautionary Tier: Why Named Candidate Genes Belong Here}

COMT deserves particular attention because it is the most mechanistically plausible candidate for a behavioral or physiological anchor, yet even here the honest conclusion is the same.

COMT encodes a methyltransferase that inactivates dopamine in the prefrontal cortex (where reuptake transporters are sparse). The Val158Met substitution (rs4680) makes the Met enzyme thermolabile---running at roughly one-third to one-quarter of Val enzyme activity---and because alleles are co-dominant, Val/Met has intermediate activity (Val/Val $>$ Val/Met $>$ Met/Met). Prefrontal dopamine follows an inverted-U performance function, so Met carriers (more dopamine) tend toward sharper baseline cognition but greater stress vulnerability, while Val carriers tend toward resilience under stress. The mechanism is real \cite{cools2011}. What is not reliable is the behavioral or physiological translation: in the autonomic domain specifically, the COMT--HRV association reverses direction with age \cite{chang2019}, and across behavioral domains the candidate-gene associations have not survived large-scale replication \cite{border2019}. The appropriate function of COMT in this framework is therefore to \emph{modulate uncertainty}---slightly raising or lowering $\sigma_G^2$ for the relevant signal---not to shift the set-point estimate $\mu_G$ confidently.

\section{From Attribution to Causation}

\begin{figure}[t]
\centering
\includegraphics[width=\textwidth]{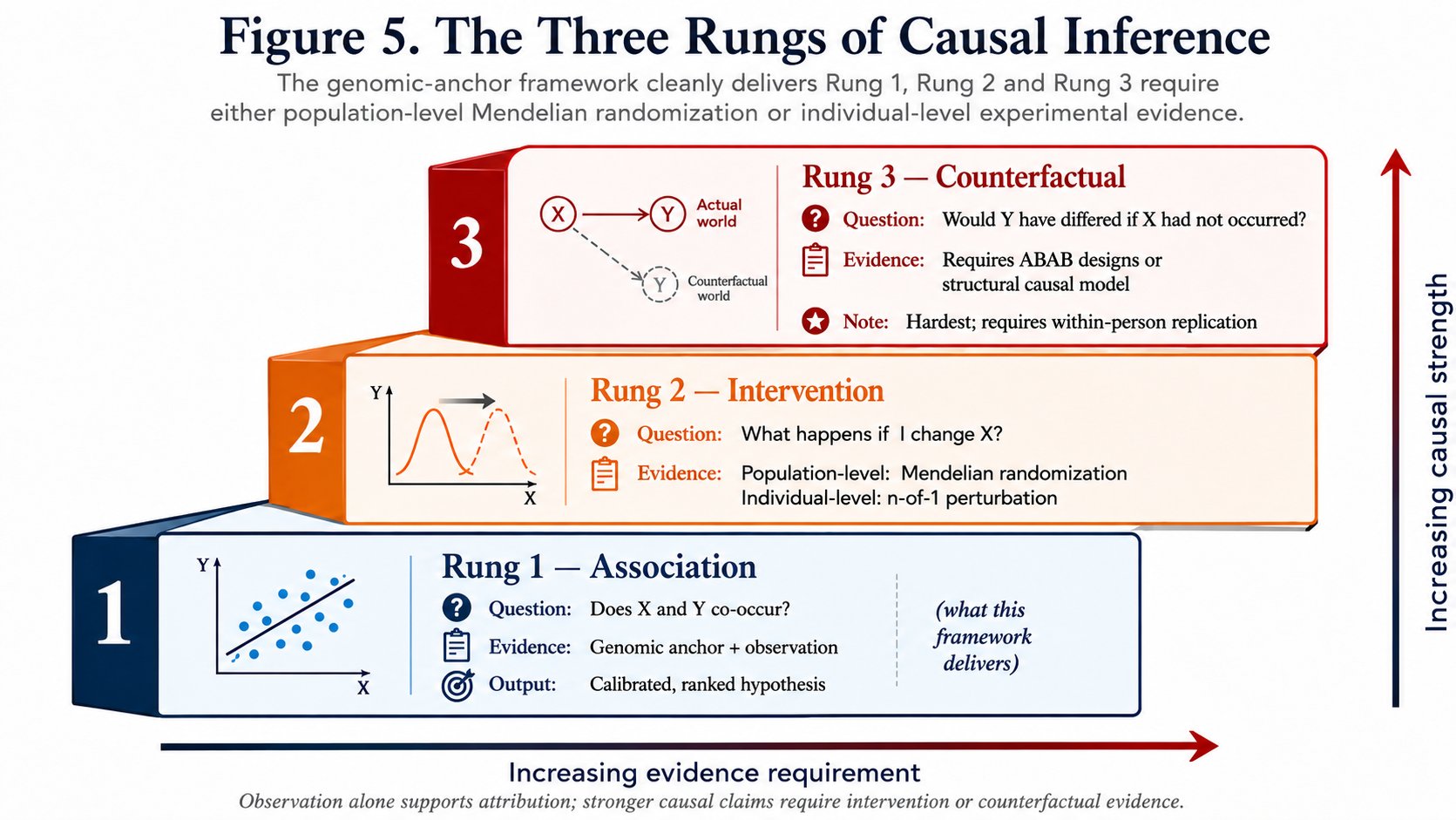}
\caption{\cite{pearl2018}. The genomic-anchor framework cleanly delivers Rung~1: a calibrated, ranked causal hypothesis. Rung~2 and Rung~3 require either population-level Mendelian randomization or individual-level experimental evidence. Observation alone supports attribution; stronger causal claims require intervention or counterfactual evidence.}
\label{fig:ladder}
\end{figure}

\subsection{Attribution: What the Exogenous Anchor Delivers}

The exogenous genetic anchor delivers causal attribution at Rung 1: because the genetic set point is exogenous, the deviation $\delta = P - \hat{G}$ is, by construction, non-genetic. Whatever is driving the deviation must lie in environment, state, or noise. This is a genuine causal gain over a population reference, because it eliminates one class of confounders---genetic constitution---by construction rather than statistical adjustment. The output is a \emph{ranked, calibrated list of candidate causes} for the deviation, not a single causal assertion.

\subsection{Population Causation: Mendelian Randomization}

At the population level, an exogenous genotype can serve as an instrumental variable in Mendelian randomization, enabling causal inference about the effect of a genetically influenced exposure on an outcome while sidestepping environmental confounders and reverse causation \cite{daveysmith2003}. For example, if FADS1/2 genotype robustly predicts circulating omega-3 levels, it can be used as an instrument to ask whether omega-3 levels causally affect inflammatory markers---a question that observational data alone cannot settle. The caveat is horizontal pleiotropy: if the instrument affects the outcome through pathways other than the exposure of interest, the causal estimate is biased \cite{bowden2015}. Behavioral and physiological traits are extensively pleiotropic \cite{burgess2019}, so MR results in this domain should be treated as suggestive.

\subsection{Individual Token Causation: The N-of-1 Path}

\begin{figure}[t]
\centering
\includegraphics[width=\textwidth]{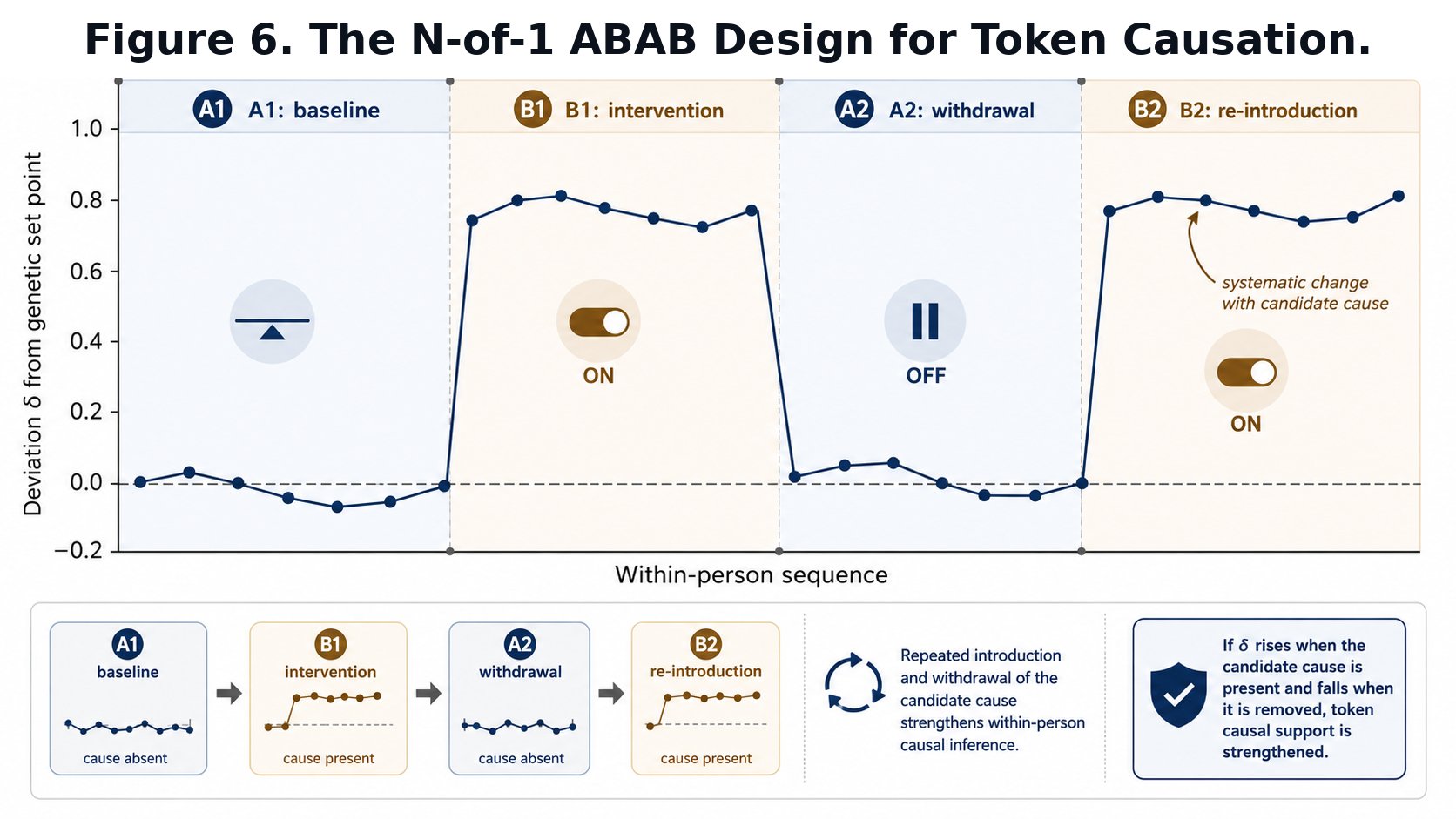}
\caption{Moving from a ranked causal hypothesis to a claim about a specific event requires within-person experimental evidence. The ABAB design---repeated introduction and withdrawal of the candidate cause---provides this \cite{schork2015}. If $\delta$ rises when the candidate cause is present and falls when it is removed, token causal support is strengthened. The exogenous genetic anchor improves power by supplying a stable constitutional reference against which perturbation effects are measured.}
\label{fig:nof1}
\end{figure}

Token causation---``did poor sleep \emph{cause} this person's HRV suppression last Tuesday''---cannot be settled by observation alone. It requires within-person experimental evidence. The n-of-1 ABAB design \cite{schork2015}, illustrated in Figure~\ref{fig:nof1}, provides this: the candidate cause is introduced and withdrawn repeatedly, and the deviation $\delta$ is tracked. If $\delta$ changes systematically with the presence and absence of the candidate, the token causal claim is supported for that individual. The exogenous genetic anchor is essential here because it provides the expected value against which changes in $\delta$ are measured---a stable, constitution-anchored reference that does not itself change with the intervention.

Absent deliberate perturbation, longitudinal temporal-precedence analysis can generate weaker but useful causal evidence: if the candidate cause systematically precedes the deviation with a plausible lag and a dose--response gradient, it constitutes a stronger causal hypothesis than a contemporaneous correlation \cite{granger1969}. But temporal precedence does not rule out an upstream common cause.

\subsection{The Honest Ceiling}

The strongest output this framework can deliver from observation plus an exogenous genetic anchor is:

\begin{center}
\begin{tikzpicture}
\node[rectangle, rounded corners=5pt, draw=signalred, fill=signalred!8, thick,
      text width=11cm, align=left, inner sep=10pt] {
\textbf{Output ceiling for a single reading:}\\[4pt]
\textit{``Based on a genetic set point of $\hat{G} = 80$\,ms HRV, the observed value of 55\,ms represents a deviation of $-25$\,ms ($-31\%$). The top-ranked non-genetic candidates for this deviation are: (1) sleep debt [strong prior evidence, testable by sleep extension over 3 nights], (2) training-load accumulation [testable by 2-day deload], (3) psychosocial stress [testable by stress diary correlation]. Confidence in genetic set-point estimate: moderate. Confidence that deviation is non-genetic: high (exogeneity). Confidence in ranking of candidates: low---requires within-person testing.''}
};
\end{tikzpicture}
\end{center}

This is attribution, not proof. It is more useful than either ``your HRV is normal'' (population) or silence (cold-start). And it is honest about where it sits on the causal ladder.

\section{The Prior-Decay Architecture}

\begin{figure}[t]
\centering
\includegraphics[width=\textwidth]{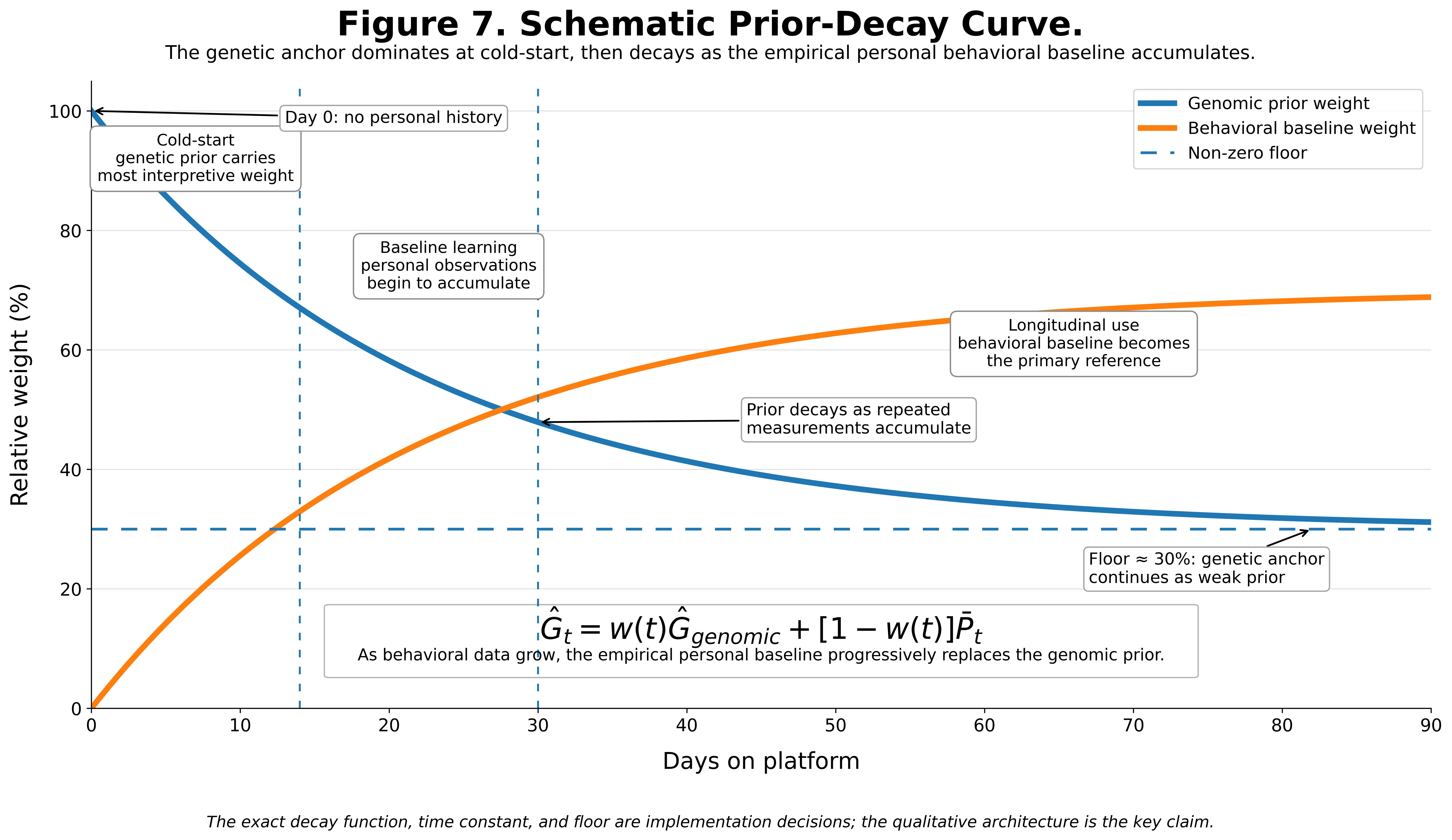}
\caption{The exogenous genetic anchor carries full weight at cold-start (Day~0) and decays as the empirical personal behavioral baseline accumulates, settling at a non-zero floor ($\approx$30\%) where it continues as a weak interpretive anchor. The blending formula $\hat{G}_t = w(t)\hat{G}_{\text{genomic}} + [1-w(t)]\bar{P}_t$ governs the transition. The exact decay function, time constant, and floor value are implementation decisions; the qualitative architecture is the key claim.}
\label{fig:decay}
\end{figure}

Because the exogenous genetic anchor is weak, it should not persist at full weight once behavioral data become available. Figure~\ref{fig:decay} illustrates the qualitative decay architecture: on Day~0, before any behavioral data exist, the exogenous genetic anchor carries the full weight for the set-point estimate. As longitudinal personal data accrue, an empirical behavioral baseline becomes estimable, and the prior's weight decays toward a non-zero floor. The floor ensures that the genetic set point continues to contribute as an interpretive reference---useful for returning from long gaps in data, for contextualising deviations from personal baseline, and for signals where behavioral data are inherently sparse. The specific decay function, time constant, and floor value are implementation decisions; the qualitative architecture is all this framework specifies.

\section{Constraints for Honest Implementation}
\label{sec:constraints}

\subsection{Weak and Evidence-Graded}

The anchor must carry trait-specific uncertainty calibrated to evidence grade (Table~\ref{tab:map}). FTO and FADS provide the strongest priors; dopaminergic and serotonergic markers provide weak directional nudges that should reduce the confidence of alerts rather than drive them.

\subsection{Ancestry-Matched}

Effect-size estimates from GWAS in one population do not transfer cleanly to another. Polygenic scores trained on predominantly European samples lose substantial predictive accuracy in South Asian, African, and East Asian populations \cite{martin2019}. A genetic set point built on mismatched betas will systematically misplace the anchor and risk compounding existing health disparities. Ancestry-matched effect sizes should be used wherever possible; reduced confidence should be flagged explicitly where only mismatched estimates are available.

\subsection{Beta Value Sourcing and the South Asian Ancestry Gap}

Every genetic set-point estimate in this framework depends on one number per SNP: the \emph{effect size} (beta value) from a GWAS, which encodes the population-average change in the trait per risk allele. Concretely, if a GWAS reports that each additional FTO risk allele is associated with a 0.36\,kg/m$^2$ increase in BMI in a European cohort, that beta---0.36---is what converts a genotype count (0, 1, or 2 risk alleles) into a set-point adjustment. The genetic set-point estimate for a given physiological signal is therefore the inner product of an individual's allele vector and the corresponding beta vector: $\hat{G} = \sum_i \beta_i g_i$, where $g_i \in \{0,1,2\}$ is the risk-allele count at locus $i$ and $\beta_i$ is the population-matched effect size.

The problem is that nearly all published betas for behavioral and physiological traits were estimated in predominantly European-ancestry cohorts. Because allele frequencies, linkage-disequilibrium structure, and gene--environment interactions differ across ancestries, these betas do not transfer reliably to other populations \cite{martin2019}. Applied to a South Asian individual, a European-derived beta may be directionally correct but quantitatively wrong---systematically over- or underestimating the genetic set point---in ways that are difficult to detect without within-population validation data.

\begin{table}[t]
\centering
\small
\renewcommand{\arraystretch}{1.4}
\begin{tabular}{>{\raggedright}p{3.2cm} >{\raggedright}p{3.5cm} >{\raggedright}p{3.0cm} >{\raggedright\arraybackslash}p{2.8cm}}
\toprule
\textbf{Domain} & \textbf{Primary GWAS resource} & \textbf{Dominant ancestry} & \textbf{South Asian availability} \\
\midrule
Body mass / appetite (FTO) & UK Biobank, GIANT consortium & European & Limited; some South Asian GWAS exist but smaller $n$ \\
Fatty-acid status (FADS1/2) & InCHIANTI, CHARGE & European & Minimal \\
Stress axis (FKBP5) & MDD/PTSD consortia & European & Very limited \\
Autonomic / HRV & UK Biobank GWAS & European & Absent \\
Circadian chronotype & UK Biobank + 23andMe & European & Absent \\
Dopaminergic / serotonergic & PGC, iPSYCH & European & Absent / contested \\
\bottomrule
\end{tabular}
\caption{\textbf{Table~3. GWAS resource coverage and South Asian availability by domain.} For most behavioral and physiological domains, population-matched beta values for South Asian individuals are unavailable, limiting the accuracy of genomic set-point estimation in this population.}
\label{tab:gwas}
\end{table}

Table~\ref{tab:gwas} maps the primary GWAS resources used for each domain to their dominant ancestry and the availability of South Asian-specific estimates. The picture is stark: for autonomic, circadian, and dopaminergic domains, no South Asian GWAS summary statistics exist at all. For metabolic domains, some smaller South Asian cohorts have been published, but sample sizes remain an order of magnitude below the European mega-analyses that anchor current effect-size estimates.

\begin{figure}[t]
\centering
\includegraphics[width=\textwidth]{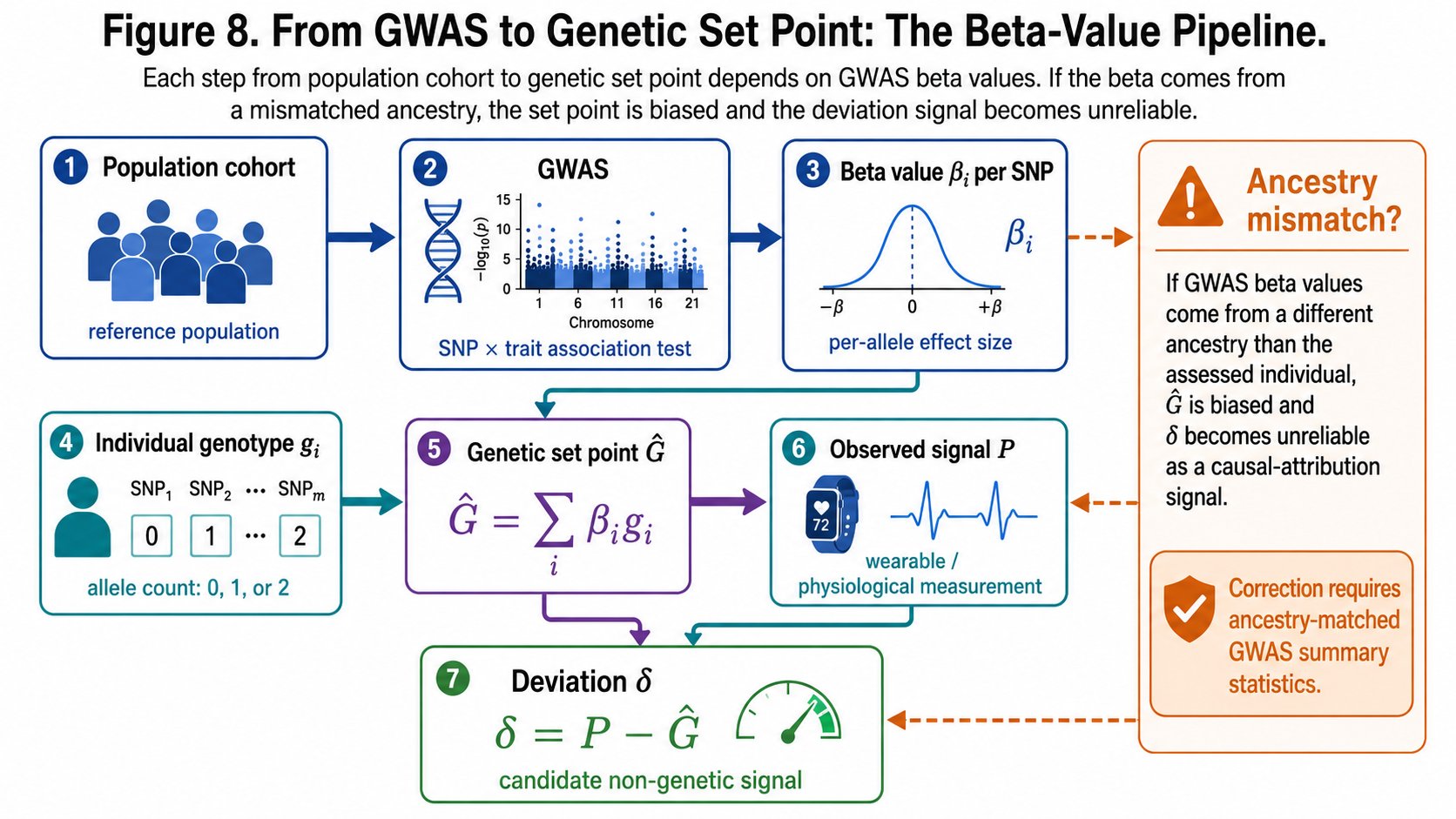}
\caption{Each step in the chain from population cohort to genetic set point $\hat{G}$ depends on
the beta value $\beta_i$---the per-allele effect size estimated by GWAS. When the beta is
derived from a different ancestry than the individual being assessed (the dominant case for
South Asian populations), the set point is biased and the deviation $\delta$ becomes
unreliable as a causal-attribution signal. Correcting this requires ancestry-matched GWAS
summary statistics.}
\label{fig:pipeline}
\end{figure}

\subsection{Dynamic and Epigenetically Aware}

The DNA sequence is fixed, but its physiological expression is not. Gene-by-environment interaction and epigenetic modulation---exemplified by FKBP5---mean that the same genotype can produce different physiological set points in different individuals depending on their environmental history. A fully dynamic anchor would incorporate epigenetic or expression-level signals, but these require periodic biological sampling (blood or saliva) rather than passive monitoring. At minimum, the framework should acknowledge that the exogenous genetic anchor captures the DNA contribution only, and that the epigenetic component may substantially shift the effective set point in individuals with strong early-adversity histories.

\subsection{Attribution, Not Diagnosis}

The framework produces ranked causal hypotheses, not diagnoses. Outputs are unsuitable for clinical diagnosis, employment decisions, insurance underwriting, or any high-stakes individual determination. The MAOA example in Table~\ref{tab:candidate} illustrates why: a weakly supported gene-by-environment association for aggression has been used in criminal-justice contexts with documented harms. The honesty of the candidate-gene tier (Table~\ref{tab:candidate}) is itself an ethical requirement, not merely a scientific one.

\section{Related Work}

The framework sits at the intersection of three research programs. \emph{Digital phenotyping} uses passive smartphone and wearable data to monitor mental and physical health states \cite{onnela2016}; the present work adds a genomic reference layer and a causal-attribution architecture to that platform. \emph{Polygenic risk scores} translate GWAS findings into individual-level genomic predictions \cite{martin2019}; the present work applies this logic to physiological set points rather than disease risk, and treats the score as a prior rather than a prediction. \emph{N-of-1 medicine} uses within-person experimental designs for personalised causal inference \cite{schork2015}; the present work formalises how an exogenous genetic anchor improves the efficiency of those designs by providing a calibrated pre-intervention reference. No prior work, to our knowledge, explicitly integrates these three programs into a unified causal-attribution framework for continuous physiological signals.

\section{Limitations and Ethical Considerations}

\paragraph{Scientific limitations.}
The recoverable genetic variance from a tractable panel is small. The decomposition ignores gene--environment correlation and interaction, both of which are non-trivial for behavioral traits. A genuinely dynamic anchor would require periodic epigenetic assays, a different data modality from passive monitoring. All set-point estimates require empirical validation against established physiological references and prospective outcomes, including calibration analysis confirming that stated confidence levels match observed accuracy.

\paragraph{Ancestry and equity.}
The genomic resources underpinning most published effect sizes were generated predominantly in European-ancestry populations. Deploying this framework in South Asian, African, or admixed populations using mismatched betas will produce systematically biased set points, disproportionately affecting populations already underserved by precision medicine. Generating ancestry-matched GWAS data---potentially using passively collected wearable behavioral phenotypes at scale---is the highest-priority scientific gap for extending this framework equitably.

\paragraph{Ethical constraints.}
Genomic and physiological inference estimates private characteristics from data that do not appear private. Informed consent, transparency, and user control are non-negotiable prerequisites. No output should be used for clinical, employment, insurance, or forensic decisions. The deterministic misreading risk of genomic information is heightened for behavioral traits, making the weak-prior discipline as much an ethical requirement as a scientific one.

\section{Conclusion}

A physiological reading is a number. A number becomes informative only against a reference. Population norms answer ``is this typical?'' but cannot say whether a person's value reflects who they constitutionally are or what their environment is currently doing to them. A personal behavioral baseline is more informative but does not exist at first contact. An exogenous genetic anchor bridges this gap: it is exogenous---fixed at conception and immune to reverse causation---and therefore a principled causal anchor from the first measurement.

Against that anchor, an observed reading decomposes into a genetic set point (``nature'') and a deviation (``nurture and state''). The deviation is the candidate-causal and actionable signal. The same HRV value carries opposite causal meanings for two people with different set points: environmental suppression in one, environmental support in the other. This is the ``normal for whom'' reversal, and it is impossible without a personalized exogenous genetic anchor.

The framework's defining discipline is calibrated restraint. The exogenous genetic anchor is genuinely informative---heritability is real---but genuinely weak, restricted to the genetic-variance slice captured by measurable common variants. It must be evidence-graded, weakest for dopaminergic and serotonergic candidate genes whose associations have largely failed replication; dynamic, decaying as behavioral data take over; ancestry-matched, with reduced confidence flagged where matched estimates are unavailable; and attributed rather than deterministic, producing ranked hypotheses rather than verdicts.

Run as a loop---exogenous genetic anchor, behavioral baseline, deviation, ranked hypotheses, within-person perturbation test, updated priors---the framework climbs the causal ladder from observation to token causation for each individual over time, without ever pretending that a single reading settled the question.

\vspace{12pt}
\noindent\rule{\textwidth}{0.4pt}

\section*{Declarations}

\paragraph{Conflict of Interest.}
Aruna Dey and Suraj Biswas are affiliated with Dots-In, an early-stage venture developing a behavioral analytics platform. The conceptual framework described in this paper is relevant to that platform. No external funding was received for this work. The authors declare no other competing financial or personal interests.

\paragraph{Funding.}
This research received no external funding. The work was conducted independently by the authors at Dots-In, an IIT Bombay incubated venture.

\paragraph{Ethics Statement.}
This paper presents a conceptual and methodological framework only. No human participants, patient data, clinical trials, or empirical datasets were collected or analysed in this study. Ethics approval was not required.

\paragraph{Author Contributions.}
Aruna Dey: conceptualization, framework design, writing (original draft). Suraj Biswas: conceptualization, technical formalization, writing (review and editing).

\paragraph{AI Use Disclosure.}
Figures~1--8 in this paper were generated with the assistance of AI image generation tools (ChatGPT / DALL·E, OpenAI). All figures were reviewed and verified by the authors for scientific accuracy and consistency with the paper's content prior to inclusion. The conceptual design, scientific framing, and content specification for each figure were provided entirely by the authors. AI assistance was used solely for visual rendering. The mathematical content, theoretical framework, and written text of this paper were authored by the human authors; large language model assistance was used for editing and LaTeX typesetting support.

\noindent\rule{\textwidth}{0.4pt}

\end{document}